\documentclass{article}
\usepackage{spconf,amsmath,graphicx}

\usepackage{amssymb}
\usepackage{url}
\usepackage{float} 
\usepackage{subfigure}
\usepackage{booktabs}
\usepackage{multirow}
\usepackage{array}

\title{Multimodal Image Fusion Based on Hybrid CNN-Transformer and Non-local Cross-modal Attention}
\name{Yu Yuan$^1$, Jiaqi Wu$^2$, Zhongliang Jing$^1$, Henry Leung$^3$, Han Pan$^1$}
\address{$^1$School of AA, Shanghai Jiao Tong University, Shanghai 200240, China\\
	$^2$School of ICE, University of Electronic Science and Technology of China, Chengdu 610097, China\\
	$^3$Department of ECE, University of Calgary, Calgary, AB T2N 1N4, Canada}
\begin{document}
%
\maketitle
\begin{abstract}
The fusion of images taken by heterogeneous sensors helps to enrich the information and improve the quality of imaging. In this article, we present a hybrid model consisting of a convolutional encoder and a Transformer-based decoder to fuse multimodal images. In the encoder, a non-local cross-modal attention block is proposed to capture both local and global dependencies of multiple source images. A branch fusion module is designed to adaptively fuse the features of the two branches. We embed a Transformer module with linear complexity in the decoder to enhance the reconstruction capability of the proposed network. Qualitative and quantitative experiments demonstrate the effectiveness of the proposed method by comparing it with existing state-of-the-art fusion models. The source code of our work is available at https://github.com/pandayuanyu/HCFusion.
\end{abstract}
\section{Introduction}

Multimodal image fusion refers to combining complementary information from multi-source images to generate a fused image of improved quality \cite{hongwaizongshu, yixuezongshu}. For example, visible images focus on the details of the scene textures, while infrared images reflect the temperature information of objects. The fusion of visible and infrared images produces an image with rich details as well as salient objects. In the field of medical imaging, the fusion of images of lesions taken by multiple instruments can contribute to a more accurate diagnosis.\\ 
\indent Feature extraction and fusion rules are two core issues in multimodal image fusion. In the past few decades, many traditional methods have been proposed to solve these two problems. 
\begin{figure}[htb]
	\centering
	\includegraphics[width=3.4in]{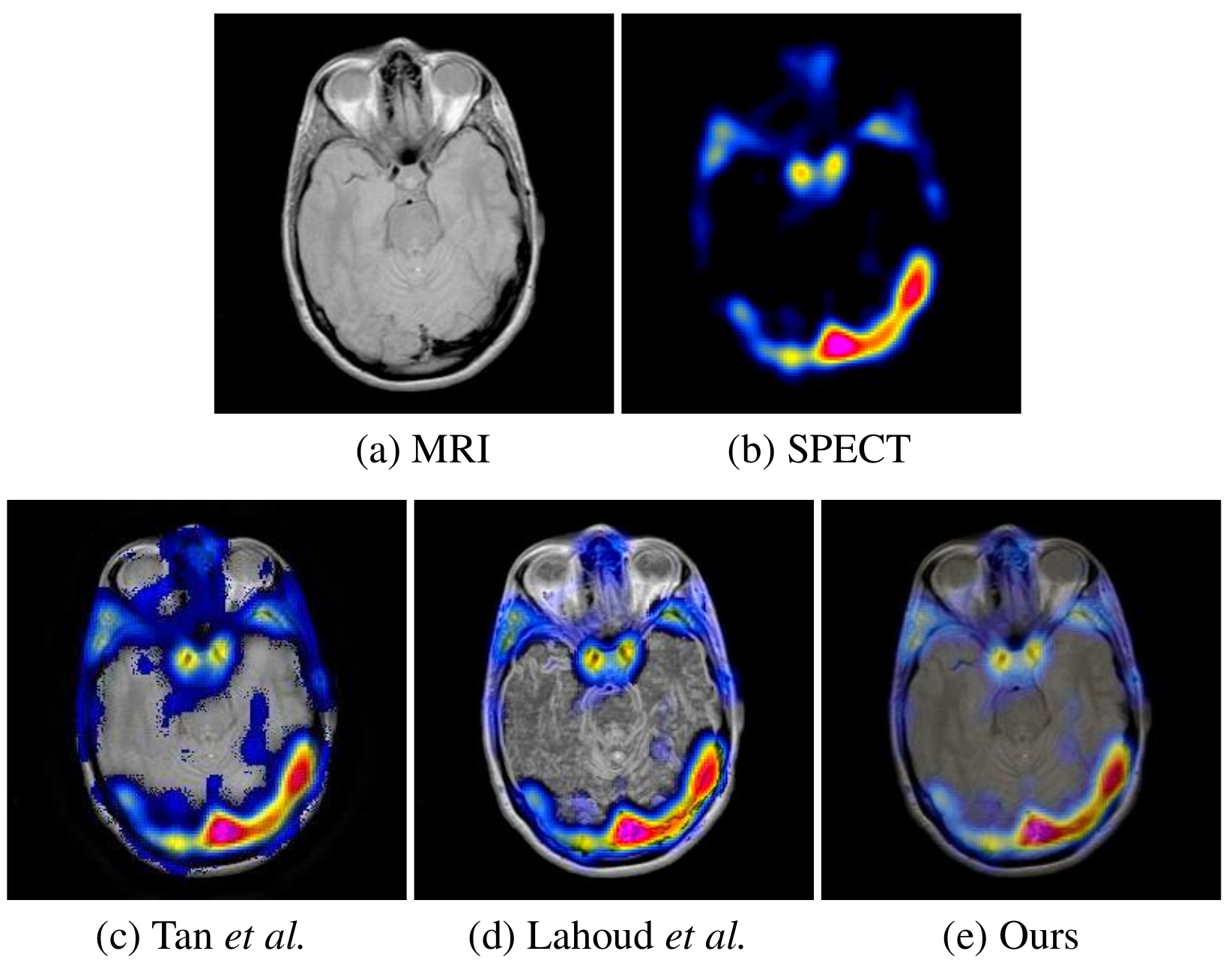}
	\caption{The fusion results for MRI and SPECT images using different methods. Tan \textit{et al.} \cite{NSST} weakens the features of SPECT. Lahoud \textit{et al.} \cite{zero} over-enhances textures of MRI. Our method generates image (e) with smoother transitions between modalities while retaining appropriate details. }
	\label{duibi_yixue}
\end{figure}
These methods can be divided into multi-scale transform \cite{multireso, NSST}, sparse representation \cite{ICASSP2022}, subspace \cite{subspace}, hybrid models \cite{hybrid}, saliency-based \cite{saliency}, and other methods \cite{hongwaizongshu}. Although these conventional methods could achieve satisfactory results in many cases, they tend to weaken the essential characteristics of source images [see Fig. \ref{duibi_yixue}(c)]. \\
\begin{figure*}[htb]
	\centering
	\includegraphics[width=7in]{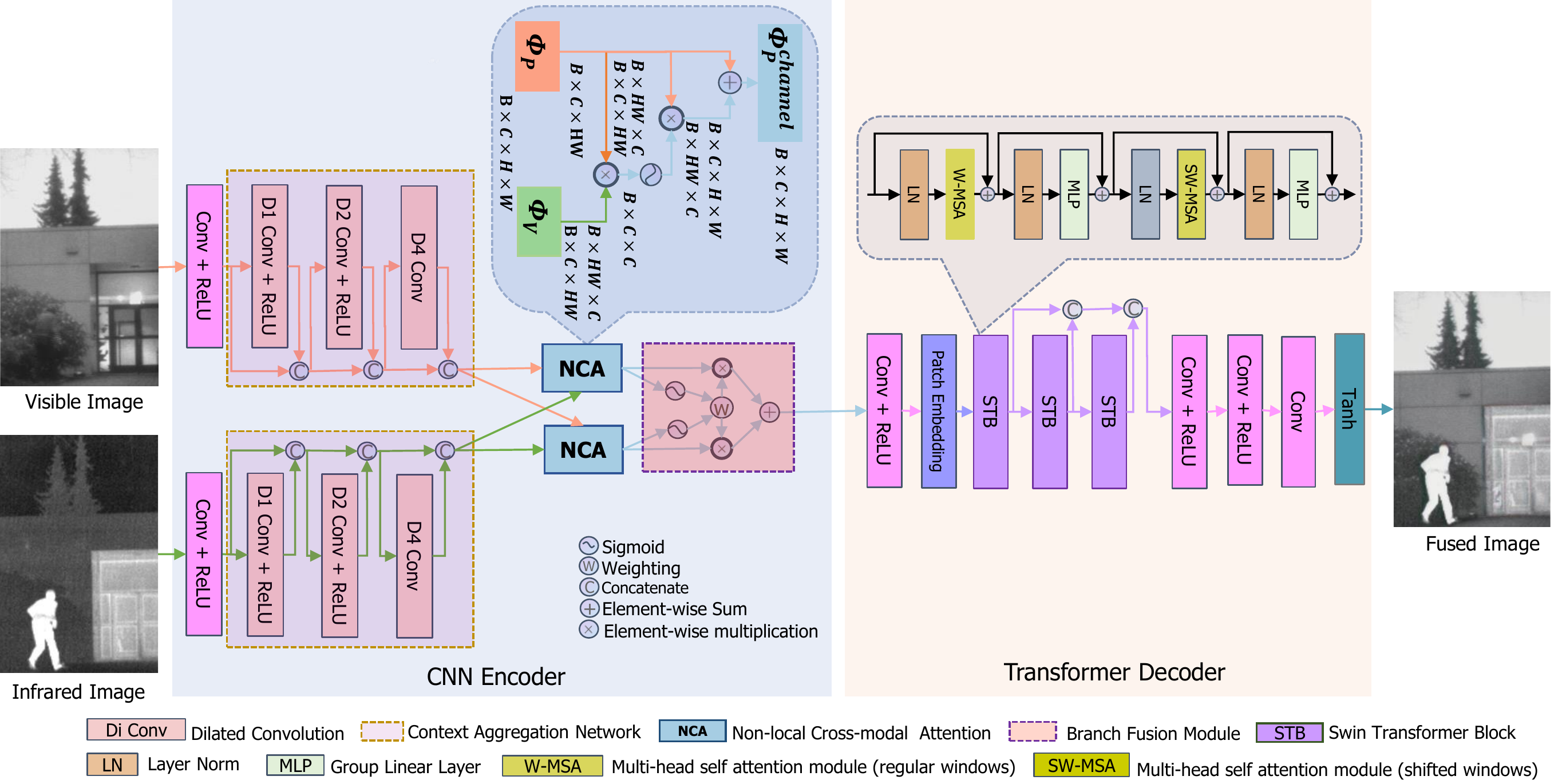}
	\caption{Proposed hybrid CNN-transformer architecture and non-local cross-modal attention block for multimodal image fusion.}
	\label{overview}
\end{figure*}
\indent Benefiting from its strong ability of feature extraction and representation, deep learning has many applications in multimodal image fusion. Prabhakar \textit{et al.} \cite{deepfuse} proposed a convolution-based network termed as Deepfuse for multi-exposure image fusion (MEF). Based on Deepfuse,  Li \textit{et al.} \cite{densefuse} proposed the visible and infrared image fusion network DenseFuse, by replacing the feature extraction layer with dense blocks and redesigning the fusion strategy. Ma \textit{et al.} \cite{fusiongan} proposed an infrared and visible image fusion framework based on a generative adversarial network (GAN). Xu \textit{et al.} \cite{UFusion} used a weight block to measure the importance of information from different source images. Lahoud \textit{et al.} \cite{zero} proposed to extract feature maps of multimodal medical images using a pre-trained network and designed a weight fusion strategy with these feature maps. Recently, several studies presented to combine fusion with high-level vision tasks. Tang \textit{et al.} \cite{SeAFusion} proposed SeAFusion, which bridged the gap between image fusion and semantic features. While these methods are valid for most fusion scenes, they emphasize the differences between different modal images without exploring their correlations. As shown in Fig. \ref{duibi_yixue}(d), the approach of Lahoud \textit{et al.} \cite{zero} enhances the features of each of the two modals, but also introduces too many trivial textures.\\
\indent In this article, we propose a network based on a hybrid CNN-Transformer architecture to fuse multimodal images. We design a non-local cross-modal attention mechanism to capture both local and global dependencies by calculating associations between features at any two locations. In addition, a branch fusion module is explored to adaptively fuse two sets of features. We conduct comparative experiments on different publicly available datasets. Qualitative and quantitative comparisons validate the superiority of our method over other conventional methods.

\section{Proposed Methods}

\subsection{Proposed Framework for Multimodal Image Fusion}
As illustrated in Fig. \ref{overview}, our network adopts a hybrid CNN-Transformer structure, which is mainly composed of a CNN-encoder and a Transformer decoder. The CNN encoder is responsible for extracting multi-scale features without losing spatial resolution by a context aggregation network (CAN) \cite{CAN}, learning the associations between the features of different modals, and performing adaptive fusion.\\
\indent The decoder is used to reconstruct fused images. Zhao \textit{et al.} \cite{hybrid2} demonstrates that the hybrid Transformer-CNN architecture has a better performance on model capacity and computational complexity. Inspired by Swin Transformer's \cite{SwinT} remarkable performance in a variety of vision tasks and the advantage of linear computational complexity, we stack three Swin Transformer blocks in the decoder to improve the reconstruction capability of the network. The patch embedding layer reduces the dimensions of the features to bridge the convolution and Transformer. It is worth noting that the application of \textit{Tanh} function in the last layer aims to constrain the range of values in the final output.

\subsection{Non-local Cross-modal Attention}
\label{ssec:subhead}

In contrast to \cite{Nonlocal, DAnet} where self-attention is used for only one input, we extend it to architecture with two inputs to obtain the connections between two different modals. As shown in Fig. \ref{overview}, we use channel attention instead of spatial attention because the multimodal data is already spatially well-aligned.

For the primary modal feature $\Phi_{P} \in \mathbb{R}^{B \times C \times H \times W}$ and the vice modal feature  $\Phi_{V} \in \mathbb{R}^{B \times C \times H \times W}$, the cross-modal channel attention operation is computed by:
\begin{equation}\label{equ41}
	y^{channel}_{i}=\frac{\sum_{\forall j} h\left(\Phi_{V_{i}}, \Phi_{P_{j}}\right) g\left(\Phi_{P_{j}}\right)}{\sum_{\forall j} h\left(\Phi_{V_{i}}, \Phi_{P_{j}}\right)}
\end{equation}
where $\Phi_{V_{i}}$ is the feature $\Phi_{V}$ at location i, and $\Phi_{P_{j}}$ represents the feature $\Phi_{P}$ at position j. Function $h$ computes the scalar which represents correlations between the two features.  Function $g$ computes the representation of the primary input at location j. $y^{channel}_{i}$ is the result of normalized aggregation of features at every position j ($\forall j$) of $\Phi_{P}$, weighted by the correlations to the secondary input $\Phi_{V}$ at location i. 

Let $Y^{channel}$ be the integration result of cross-modal channel attention operators at all position i. Then the final output of the proposed NCA is:

\begin{equation}\label{equ5}
	\Phi_{P}^{channel}=\Phi_{P}+\alpha Y^{channel}
\end{equation}
where $\alpha$ is a learnable parameter.

\subsection{Branch Fusion Module}
\label{ssec:subhead}

To fuse the two features $\Phi_{k}^{channel}(k=1,2)$ (k denotes the branch of different modal features) obtained by the proposed non-local cross-model attention block, we design a weighted branch fusion strategy. The weight of branch k is calculated by: 

\begin{equation}\label{equ1}
	w_{\Phi_{1}^{channel}}=\frac{\operatorname{sigmoid}\left(\Phi_{1}^{channel}\right)}{\operatorname{sigmoid}\left(\Phi_{1}^{channel}\right)+\operatorname{sigmoid}\left(\Phi_{2}^{channel}\right)+\varepsilon}
\end{equation}

\begin{equation}\label{equ2}
	w_{\Phi_{2}^{channel}}=\frac{\operatorname{sigmoid}\left(\Phi_{2}^{channel}\right)}{\operatorname{sigmoid}\left(\Phi_{1}^{channel}\right)+\operatorname{sigmoid}\left(\Phi_{2}^{channel}\right)+\varepsilon}
\end{equation}
where the $\text { sigmoid }$  function maps the features data into the range of (0,1), and $\varepsilon$ is set as $1e-8$. Then  $\Phi_{1}^{channel}$ and $\Phi_{2}^{channel}$ are summed according to their weights:

\begin{equation}\label{equ3}
	\Phi^{f}=w_{\Phi_{1}^{channel}} \odot \Phi_{1}^{channel}+w_{\Phi_{2}^{channel}} \odot \Phi_{2}^{channel}
\end{equation}
where $\odot$ denotes Hadamard product. This fused feature $\Phi^{f}$ is then fed into the Transformer-based decoder.

\subsection{Training Strategy and Loss Function}

Similar to \cite{densefuse}, we feed the same image to both modal branches during training. Our network mainly learns how to extract features and reconstruct the fused image at this stage. This unsupervised approach makes our network independent of specific datasets and ground truth, and thus can be applied to the fusion of multiple modal images. 
SSIM \cite{SSIM} loss and MSE loss are calculated from the input and output.\\
\begin{equation}\label{equ14}
	\begin{aligned}
		L_{MSE}&=\|\mathbf{Output}-  \mathbf{Input}\|_{2}\\
		L_{SSIM}&=1 - S S I M(\mathbf{Output}, \mathbf{Input})
	\end{aligned}
\end{equation}

Eventually, we use the following total loss function in our approach.
\begin{equation}\label{equ15}
	L_{\text {total }}=L_{MSE}+\lambda_{1} L_{SSIM}
\end{equation}
\indent The \textit{AVA} \cite{AVA} dataset is selected to train our network since it contains RGB images and grayscale images. The size of images fed into our model is $256 \times 256$, and the batch size is set to 16 for 8 epochs. $ \lambda_{1}$ is set as 10. The initial learning rate is set as $1e-4$ and decreases to $1e-8$ with the cosine annealing strategy. The AdamW \cite{AdamW} optimizer is adopted in our network.  We fix random seeds to make the ablation experiment more reliable. All experiments are performed on NVIDIA Geforce RTX 3090 GPU and Intel Core i9-10900k CPU @ 3.70GHz. Our network is programmed on PyTorch.
\begin{table}[H]
	\caption{\textbf{Average metric values of ablation experiments, on the VIFB dataset (10 image pairs). Values in bold indicate the best results.}}
	\label{ablation}
	\centering
	\setlength{\tabcolsep}{1.8mm}{
		\begin{tabular}{l|ccc}
			\toprule
			&   PSNR$\uparrow$  &  FMI$\uparrow$ & $\mathrm{Q}_{cv}$$\downarrow$     \\
			\midrule
			W/o STB        & 57.8932   &     1.9630    &   535.3597      \\
			W/o NCA   &   56.9978  &  1.7683      &   639.3315           \\
			W/o BFM   &  57.8641    &	1.8585        &  573.3622        \\
			\midrule
			Ours       &\textbf{58.4304} &\textbf{2.0060}&\textbf{520.3854} \\
			\bottomrule
	\end{tabular}}
\end{table}
\begin{figure}[htb]
	\centering
	\includegraphics[width=3.4in]{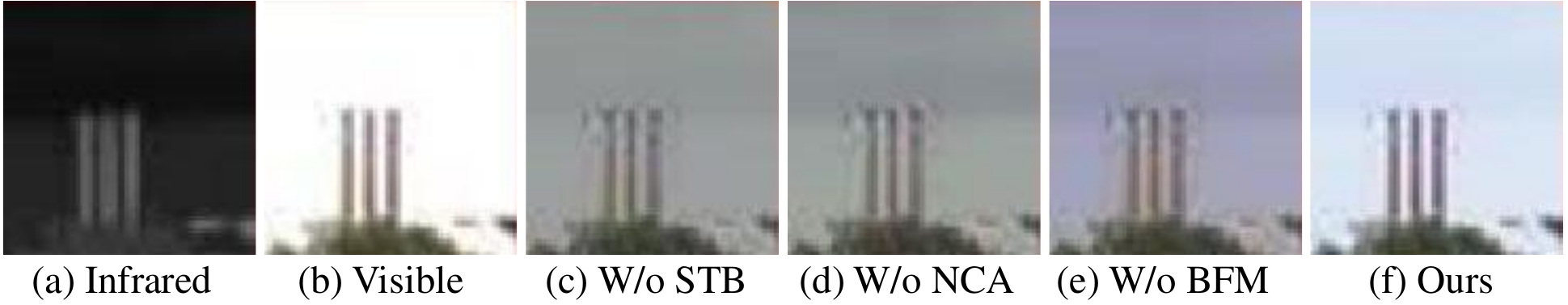}
	\caption{Local enlargements of ablation experiments on the \textit{carWhite} image in \textit{VIFB} dataset.}
	\label{xiaorongtu}
\end{figure}
\section{Experimental Results}

\label{sec:print}
We select the \textit{TNO} \cite{TNO} and \textit{VIFB} \cite{VIFB} datasets to evaluate the visible and infrared image fusion performance, using the data from \cite{haver} to evaluate the multimodal medical image fusion effect. Three metrics are utilized to quantify the fusion results, including peak signal-to-noise ratio (PSNR) \cite{PSNR}, feature mutual information (FMI) \cite{PSNR}, and human visual perception ($\mathrm{Q}_{cv}$) \cite{Qcv}.
\subsection{Ablation Experiments}
\begin{figure*}[htb]
	\centering
	\includegraphics[width=7in]{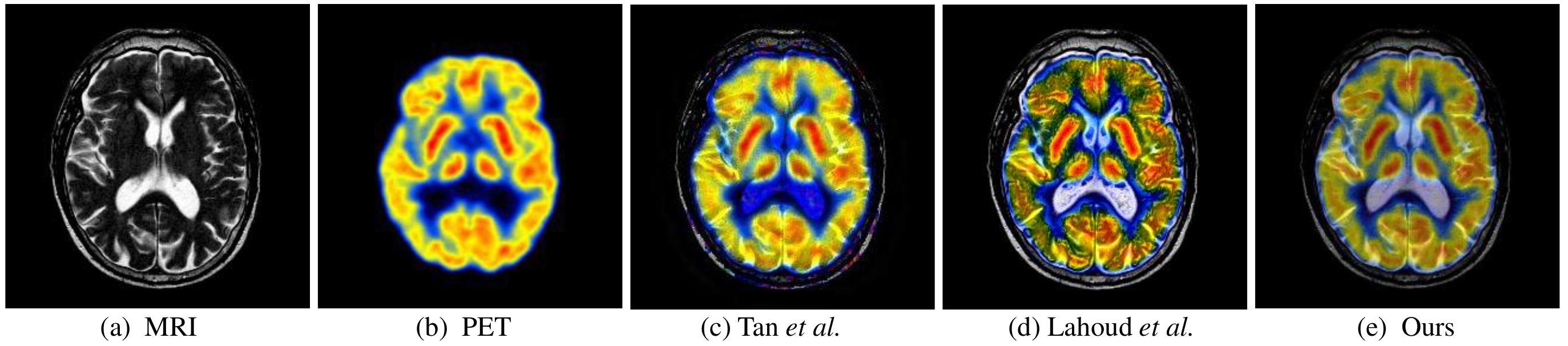}
	\caption{The fusion results for MRI and PET images using different methods.}
	\label{duibi_yixue2}
\end{figure*}
\begin{figure*}[htb]
	\centering
	\includegraphics[width=7in]{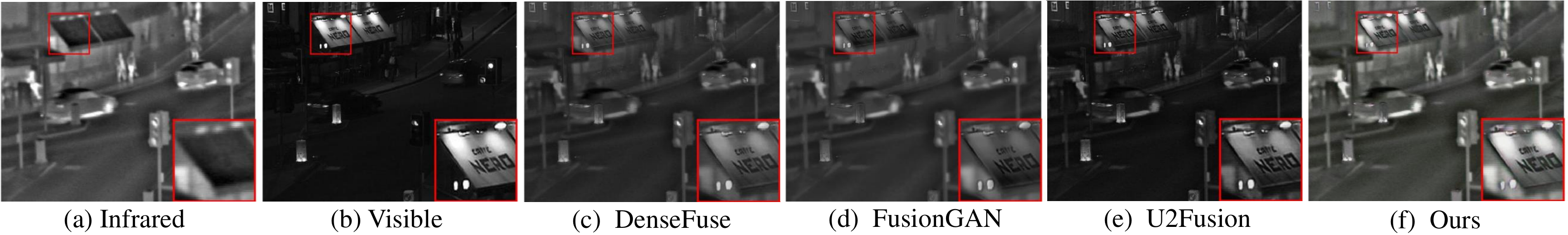}
	\caption{The fusion results for infrared and visible images using different methods.}
	\label{duibi_hongwai}
\end{figure*}
\indent {\textit{Swin Transformer Blocks:}} To explore the contribution of the embedded three consecutive Swin Transformer Blocks (STB), they are replaced by convolution and activation layers with the same dimensions. As shown in Table \ref{ablation}, the hybrid CNN-Transformer scheme outperforms the pure CNN scheme by about 0.54db in PSNR, and leads in terms of FMI and  $\mathrm{Q}_{cv}$.  Fig. \ref{xiaorongtu} shows that our method produces images with clearer textures and higher contrast.\\
\indent {\textit{Non-local Cross-modal Attention:}} We abandon the proposed non-local  cross-modal attention (NCA) module to verify its impact. Table \ref{ablation} demonstrates that the presence of NCA can effectively improve the quality of the fused images. As exhibited in Fig. \ref{xiaorongtu}, the introduction of NCA can significantly improve the details of the fused image.\\
\indent {\textit{Branch Fusion Modules:}} We remove the branch fusion module (BFM) and sum the features of two main branches on average to evaluate the effect of BFM. From Table \ref{ablation}, the model without BFM shows a decrease in all indicators.  In Fig. \ref{xiaorongtu}, it can be seen that the sharpness of the image decreases after removing the BFM.
\begin{table}[!ht]
	\centering
	\caption{\textbf{Average metric values with different multimodal fusion methods. 4 pairs of images from \cite{haver} for Medical. 42 pairs of images from  \textit{TNO} \cite{TNO} for visible and infrared (VI-IR). Values in bold indicate the best results. }}
	\label{dabijiao}
	\setlength{\tabcolsep}{1.6mm}{
		\begin{tabular}{ll|lll}
			\toprule
			& Methods & PSNR$\uparrow$ & FMI$\uparrow$  & $\mathrm{Q}_{cv}$$\downarrow$  \\ \hline
			\multirow{3}*{Medical}  & Tan \textit{et al.} & 55.6885 & 0.8246 & 1456.5711  \\
			~ & Lahoud \textit{et al.} & 54.8438 & 0.8362 & 1648.6924\\
			~ & Ours & \textbf{56.3492} & \textbf{0.8432} & \textbf{953.1129}  \\ \hline
			\multirow{4}*{VI-IR} & DenseFuse & 58.4742 & 2.0383 & 548.9475 \\
			~ & FusionGAN & 57.7786 & \textbf{2.1437} & 1063.2053  \\
			~ & U2Fusion & 58.6001 & 1.9800 & 579.4094   \\ 
			~ & Ours & \textbf{59.1492} & 2.0925 & \textbf{457.7126}  \\
			\bottomrule
	\end{tabular}}
\end{table}

\subsection{Comparative Study}
The comparative study is performed with two medical multimodal fusion methods, including a multi-scale transform-based methods (Tan \textit{et al.} \cite{NSST}) and a deep learning-based method (Lahoud \textit{et al.} \cite{zero}). For visible and infrared image fusion, we compare three networks: DenseFuse \cite{densefuse}, FusionGAN \cite{fusiongan}, and U2Fusion \cite{UFusion}.\\
\indent From Table \ref{dabijiao}, our method achieves the best performance on PSNR and $\mathrm{Q}_{cv}$, both for Medical and VI-IR. The excellent performance on $\mathrm{Q}_{cv}$ illustrates that the images produced by our method are more in line with human visual perception. From Fig. \ref{duibi_yixue2}, the method of Tan \textit{et al.} leads to a low weight of MRI image, while our approach preserves these elements well. The image generated by Tan \textit{et al.}  introduces an excessive amount of details, while our method reconciles the details of the two modalities. As exhibited in Fig. \ref{duibi_hongwai},  DenseFuse and FusionGAN fail to preserve the visible parts of the board. U2Fusion has a lower ability to highlight salient targets from infrared images.  In general, the images achieved by our method have the best visual performance.

\section{Conclusions}
\label{sec:print}

In this study, we propose a hybrid CNN-Transformer architecture by developing a non-local cross-modal attention mechanism for multimodal image fusion. We use the proposed non-local cross-modal attention to learn associations between heterologous features and then perform adaptive branch fusion. The introduction of Swin Transformer blocks improves the performance of our network. It is worth noting that our model only needs ordinary datasets rather than well-registered multimodal datasets for training. Extensive experiments demonstrate the superiority of our proposed method over existing state-of-the-art fusion methods, both objectively and subjectively.

\vfill\pagebreak

\bibliographystyle{IEEEbib}
\bibliography{strings}

\begin{thebibliography}{10}

\bibitem{hongwaizongshu}
J.~Ma, Y.~Ma, and C.~Li,
\newblock ``Infrared and visible image fusion methods and applications: A
  survey,''
\newblock {\em Information Fusion}, vol. 45, pp. 153--178, 2019.

\bibitem{yixuezongshu}
B.~Huang, F.~Yang, M.~Yin, X.~Mo, and C.~Zhong,
\newblock ``A review of multimodal medical image fusion techniques,''
\newblock {\em Computational and Mathematical Methods in Medicine}, vol. 2020,
  pp. 1--16, 04 2020.

\bibitem{multireso}
S.~Li, B.~Yang, and J.~Hu,
\newblock ``Performance comparison of different multi-resolution transforms for
  image fusion,''
\newblock {\em Information Fusion}, vol. 12, no. 2, pp. 74--84, 2011.

\bibitem{NSST}
W.~Tan, P.~Tiwari, H.~M. Pandey, C.~Moreira, and A.~Jaiswal,
\newblock ``Multimodal medical image fusion algorithm in the era of big data,''
\newblock {\em Neural Computing and Applications}, pp. 1--21, 07 2020.

\bibitem{ICASSP2022}
F.~G. Veshki and S.~A. Vorobyo,
\newblock ``Coupled feature learning via structured convolutional sparse coding
  for multimodal image fusion,''
\newblock in {\em ICASSP}, 2022, pp. 2500--2504.

\bibitem{subspace}
D.~P. Bavirisetti, G.~Xiao, and G.~Liu,
\newblock ``Multi-sensor image fusion based on fourth order partial
  differential equations,''
\newblock in {\em ICIF}, 2017, pp. 1--9.

\bibitem{hybrid}
J.~Zhao, Y.~Chen, H.~Feng, Z.~Xu, and Q.~Li,
\newblock ``Infrared image enhancement through saliency feature analysis based
  on multi-scale decomposition,''
\newblock {\em Infrared Physics and Technology}, vol. 62, pp. 86--93, 2014.

\bibitem{saliency}
D.~P. Bavirisetti and R.~Dhuli,
\newblock ``Two-scale image fusion of visible and infrared images using
  saliency detection,''
\newblock {\em Infrared Physics and Technology}, vol. 76, pp. 52--64, 2016.

\bibitem{deepfuse}
K.~R. Prabhakar, V.~S. Srikar, and R.~V. Babu,
\newblock ``Deepfuse: A deep unsupervised approach for exposure fusion with
  extreme exposure image pairs,''
\newblock in {\em ICCV}, 2017, pp. 4724--4732.

\bibitem{densefuse}
H.~Li and X.~Wu,
\newblock ``Densefuse: A fusion approach to infrared and visible images,''
\newblock {\em IEEE Transactions on Image Processing}, vol. 28, no. 5, pp.
  2614--2623, 2019.

\bibitem{fusiongan}
J.~Ma, W.~Yu, P.~Liang, C.~Li, and J.~Jiang,
\newblock ``Fusiongan: A generative adversarial network for infrared and
  visible image fusion,''
\newblock {\em Information Fusion}, vol. 48, pp. 11--26, 2019.

\bibitem{UFusion}
H.~Xu, J.~Ma, J.~Jiang, X.~Guo, and H.~Ling,
\newblock ``U2fusion: A unified unsupervised image fusion network,''
\newblock {\em IEEE Transactions on Pattern Analysis and Machine Intelligence},
  vol. 44, no. 1, pp. 502--518, 2022.

\bibitem{zero}
F.~Lahoud and S.~Süsstrunk,
\newblock ``Zero-learning fast medical image fusion,''
\newblock in {\em FUSION}, 2019, pp. 1--8.

\bibitem{SeAFusion}
L.~Tang, J.~Yuan, and J.~Ma,
\newblock ``Image fusion in the loop of high-level vision tasks: A
  semantic-aware real-time infrared and visible image fusion network,''
\newblock {\em Information Fusion}, vol. 82, pp. 28--42, 2022.

\bibitem{CAN}
F.~Yu and V.~Koltun,
\newblock ``Multi-scale context aggregation by dilated convolutions,''
\newblock {\em CoRR}, vol. abs/1511.07122, 2016.

\bibitem{hybrid2}
M.~Zhao, G.~Cao, X.~Huang, and L.~Yang,
\newblock ``Hybrid transformer-cnn for real image denoising,''
\newblock {\em IEEE Signal Processing Letters}, vol. 29, pp. 1252--1256, 2022.

\bibitem{SwinT}
Z.~Liu, Y.~Lin, Y.~Cao, H.~Hu, Y.~Wei, Z.~Zhang, S.~Lin, and B.~Guo,
\newblock ``Swin transformer: Hierarchical vision transformer using shifted
  windows,''
\newblock in {\em ICCV}, 2021, pp. 9992--10002.

\bibitem{Nonlocal}
X.~Wang, R.~Girshick, A.~Gupta, and K.~He,
\newblock ``Non-local neural networks,''
\newblock in {\em CVPR}, 2018, pp. 7794--7803.

\bibitem{DAnet}
J.~Fu, J.~Liu, H.~Tian, Y.~Li, Y.~Bao, Z.~Fang, and H.~Lu,
\newblock ``Dual attention network for scene segmentation,''
\newblock in {\em CVPR}, 2019, pp. 3141--3149.

\bibitem{SSIM}
Z.~Wang, A.~C. Bovik, H.~R. Sheikh, and E.~P. Simoncelli,
\newblock ``Image quality assessment: from error visibility to structural
  similarity,''
\newblock {\em IEEE Transactions on Image Processing}, vol. 13, no. 4, pp.
  600--612, 2004.

\bibitem{AVA}
N.~Murray, L.~Marchesotti, and F.~Perronnin,
\newblock ``Ava: A large-scale database for aesthetic visual analysis,''
\newblock in {\em 2012 IEEE Conference on Computer Vision and Pattern
  Recognition}, 2012, pp. 2408--2415.

\bibitem{AdamW}
I.~Loshchilov and F.~Hutter,
\newblock ``Decoupled weight decay regularization,''
\newblock in {\em ICLR}, 2019.

\bibitem{TNO}
{\em \url{https://figshare.com/articles/dataset/TNO_Image_Fusion_Dataset}}.

\bibitem{VIFB}
X.~Zhang, P.~Ye, and G.~Xiao,
\newblock ``Vifb: A visible and infrared image fusion benchmark,''
\newblock in {\em CVPRW}, 2020, pp. 468--478.

\bibitem{haver}
{\em \url{https://www.med.harvard.edu/aanlib}}.

\bibitem{PSNR}
P.~Jagalingam and A.~V. Hegde,
\newblock ``A review of quality metrics for fused image,''
\newblock {\em Aquatic Procedia}, vol. 4, pp. 133--142, 2015.

\bibitem{Qcv}
H.~Chen and P.~K. Varshney,
\newblock ``A human perception inspired quality metric for image fusion based
  on regional information,''
\newblock {\em Information Fusion}, vol. 8, no. 2, pp. 193--207, 2007.

\end{thebibliography}

\end{document}